\documentclass{article}

\usepackage{arxiv}

\UseRawInputEncoding
\usepackage[T1]{fontenc}    
\usepackage{hyperref}       
\usepackage{url}            
\usepackage{booktabs}       
\usepackage{amsfonts}       
\usepackage{nicefrac}       
\usepackage{microtype}      
\usepackage{lipsum}		
\usepackage{graphicx}
\usepackage[numbers]{natbib}
\usepackage{doi}
\usepackage{dsfont}
\usepackage{amsmath}
\usepackage{xcolor}

\newcommand{\IoUmod}{\mathrm{IoU}^{\mathrm{(modified)}}}
\newcommand{\IoUh}{\mathrm{IoU}^{\mathrm{(match)}}}
\newcommand{\IoUl}{\mathrm{IoU}^{\mathrm{(softmatch)}}}
\newcommand{\SoftPQlh}{\mathrm{SoftPQ}(l,h)} 
\newcommand{\GTcount}{|G|} 
\newcommand{\nsoft}{n}
\newcommand{\nmatch}{m}
\newcommand{\softpq}{\textbf{SoftPQ}}
\newcommand{\papertitle}{SoftPQ: Robust Instance Segmentation Evaluation via Soft Matching and Tunable Thresholds}

\title{\papertitle}

\author{ \href{https://orcid.org/0000-0003-2427-7920}{\includegraphics[scale=0.06]{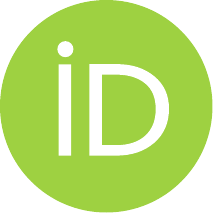}\hspace{1mm}Ranit Karmakar}\thanks{Corresponding author} \\
	Department of Systems Biology\\
	Harvard Medical School\\
	Boston, MA 02115 \\
	\texttt{ranit@hms.harvard.edu} \\
	\And
	\href{https://orcid.org/0000-0001-8302-526X}{\includegraphics[scale=0.06]{orcid.pdf}\hspace{1mm}Simon F. N{\o}rrelykke} \\
	Department of Systems Biology\\
	Harvard Medical School\\
	Boston, MA 02115 \\
	\texttt{simon@hms.harvard.edu}
}

\renewcommand{\shorttitle}{\papertitle}

\begin{document}
\maketitle

\begin{abstract}
Segmentation evaluation metrics traditionally rely on binary decision logic: predictions are either correct or incorrect, based on rigid IoU thresholds. Detection-based metrics such as F1 and mAP determine correctness at the object level using fixed overlap cutoffs, while overlap-based metrics like Intersection over Union (IoU) and Dice operate at the pixel level, often overlooking instance-level structure. Panoptic Quality (PQ) attempts to unify detection and segmentation assessment, but it remains dependent on hard-threshold matching—treating predictions below the threshold as entirely incorrect. This binary framing obscures important distinctions between qualitatively different errors and fails to reward gradual model improvements. We propose \softpq\, a flexible and interpretable instance segmentation metric that redefines evaluation as a graded continuum rather than a binary classification. \softpq\ introduces tunable upper and lower IoU thresholds to define a partial matching region and applies a sublinear penalty function to ambiguous or fragmented predictions. These extensions allow \softpq\ to exhibit smoother score behavior, greater robustness to structural segmentation errors, and more informative feedback for model development and evaluation. Through controlled perturbation experiments, we show that \softpq\ captures meaningful differences in segmentation quality that existing metrics overlook, making it a practical and principled alternative for both benchmarking and iterative model refinement.
\end{abstract}


\section{Introduction}
\label{sec:intro}
Evaluating instance segmentation performance remains a core task in computer vision. Standard metrics such as the F1 score, Intersection over Union (IoU), mean Average Precision (mAP), and Panoptic Quality (PQ) are widely used to assess segmentation quality. While these metrics are effective for benchmarking, they tend to be highly sensitive to minor prediction (segmentation) discrepancies. While useful for rigor, this sensitivity can obscure improvements and yield counterintuitive results during development.

\begin{figure}
    \label{fig:limitations}
  \centering
  \includegraphics[width=0.99\linewidth]{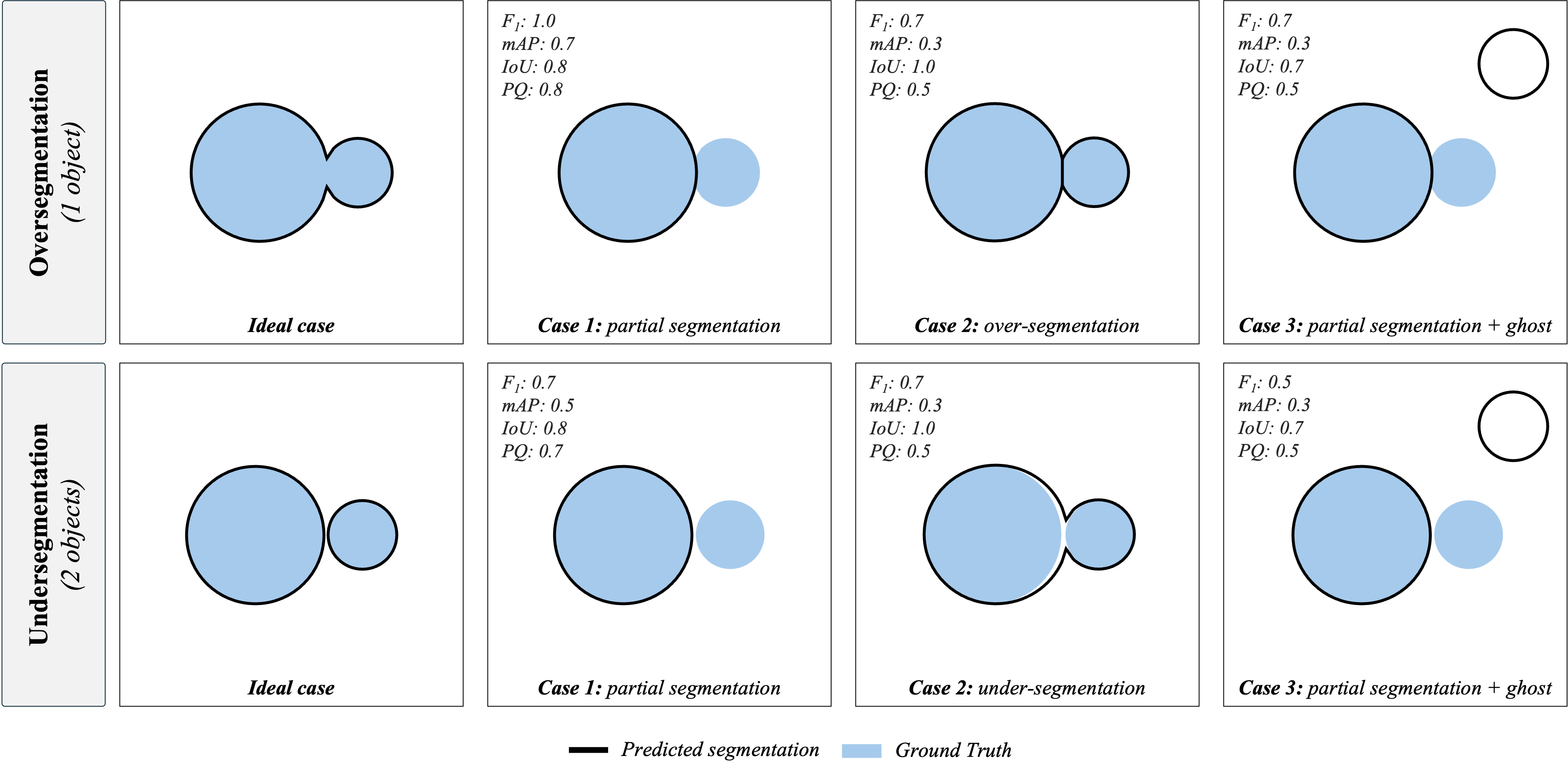}
  \caption{\textbf{Limitations of Standard Segmentation Metrics in Handling over- and under-segmentation:} Synthetic examples illustrating common segmentation errors—partial segmentation, over-segmentation, and ghost segments—for both single-object and multi-object cases. Despite there being qualitative differences in the predicted masks, traditional metrics (F1, mAP, IoU, PQ) often produce same or similar scores, highlighting their limited ability to distinguish between qualitatively distinct errors.}
\end{figure}

As illustrated in Figure~\ref{fig:limitations}, predicted masks with visibly and qualitatively different segmentation errors can yield similar scores when using conventional metrics. This inability to differentiate between error types makes it difficult to trust these metrics for diagnostic or iterative model improvement. Moreover, small changes in the predicted mask can lead to large variations in metric scores, undermining their interpretability and stability.

To address these limitations, we introduce \textbf{Soft Panoptic Quality (\softpq)}, a principled extension and generalization of the PQ metric \cite{kirillov2019panoptic}. \softpq\ incorporates two tunable IoU thresholds—an upper and a lower bound—that define a range for valid matches. The upper threshold identifies strong matches between predicted and ground truth segments, similar to the original PQ formulation. The lower threshold captures "soft" matches associated with over- and under-segmentation, allowing for a more nuanced interpretation of partial overlaps. Users can tune these parameters based on task specific sensitivity. Notably, setting both thresholds to 0.5 recovers the original PQ as a special case of \softpq.

In addition to threshold tuning, \softpq\ modifies how match quality is aggregated. It introduces a sublinear penalty to ambiguous matches by applying a normalization factor of $\sqrt{n+1}$ where $n$ is the number of partially matched segments. This penalizes soft matches while still reflecting gradual improvements in segmentation quality.

We evaluate \softpq\ using a series of controlled synthetic experiments designed to illustrate common segmentation failure modes, including over- and under-segmentation. Our results show that \softpq\ provides a more robust and interpretable assessment across diverse error types. Compared to traditional metrics, it better captures the qualitative differences between outputs, offering a more useful signal for model evaluation and development.

\section{Related Work}
\label{sec:literature}
Segmentation evaluation metrics are commonly categorized into detection-based and segmentation-based types. Detection-based metrics, such as the F1 score~\cite{taha2015metrics} and mAP~\cite{lin2014microsoft}, assess how many true labels are accurately detected based on threshold matching. mAP also allows different threshold levels between $0.5$ and $0.9$ for controlled scoring. In contrast, segmentation-based metrics like IoU~\cite{taha2015metrics} and Dice coefficient~\cite{taha2015metrics} focus on the degree of pixel-level overlap between predicted and ground truth masks. Each class of metric captures a different aspect of performance and has its limitations when used in isolation.

To bridge this gap, Panoptic Quality (PQ)\cite{kirillov2019panoptic} was introduced as a unified metric that combines detection accuracy (via recognition quality, RQ) and segmentation quality (SQ) into a single score. Originally proposed for panoptic segmentation, PQ has since been adopted in instance segmentation as a balanced metric that accounts for both instance-level correctness and mask quality\cite{maier2024metrics}. However, its reliance on a fixed IoU threshold for matching predictions with ground truth makes it sensitive to small changes in mask shape or placement, which can lead to counterintuitive score drops in edge cases involving over-segmentation, under-segmentation, or ambiguous boundaries.

To mitigate such issues, several alternative strategies have been proposed. Boundary IoU~\cite{cheng2021boundaryiou} emphasizes alignment of object contours rather than overall region overlap, improving relevance in settings with complex or fine-grained structures. SortedAP~\cite{chen2023sortedap} redefines average precision using instance-wise rankings to promote more stable and interpretable evaluations. In specialized domains like medical imaging, metric adaptations often incorporate task-specific priors or modify loss functions to better reflect domain-relevant errors~\cite{zhou2022boundaryloss, blaschko2023metrics}.

Recent work has also challenged the suitability of PQ in biomedical contexts, where the assumption of hard thresholds may fail to reflect gradual or partial improvements in segmentation quality~\cite{jena2023beyondmap, foucart2023pqshouldbeavoided}, a critical point when fine-tuning an image analysis pipeline. Broader reviews of segmentation metrics~\cite{liu2022review, zhou2023advances} highlight the need for more flexible and interpretable metrics that are sensitive to context-specific requirements and qualitative performance differences.

In this setting, we propose \softpq\, which extends PQ by introducing tunable IoU thresholds and a sublinear aggregation function to handle partial matches. This formulation maintains the attractive features of PQ while offering a more stable and adjustable evaluation of segmentation performance. \softpq\ is designed to support both fine-grained analysis during model development and consistent benchmarking in diverse real-world applications.

\section{Proposed Method: Soft Panoptic Segmentation with \softpq}
\label{sec:method}
We propose \softpq\, an extension of the PQ metric, designed to provide a more flexible and interpretable framework for evaluating segmentation quality. Unlike the standard PQ metric, which uses a fixed IoU threshold to determine matching between predicted and ground truth segments, \softpq\ introduces two tunable parameters:

\begin{itemize}
    \item \textbf{Upper IoU threshold ($h$):} Defines the minimum IoU required for a segment to be considered a full match, i.e., a true positive.
    \item \textbf{Lower IoU threshold ($l$):} Allows partial matches (soft-matching) to be incorporated into the score in case of over- and under-segmentation.
\end{itemize}

This dual-threshold formulation relaxes the binary matching assumption in traditional PQ and provides a more continuous assessment of segmentation quality. Setting $h > 0.5$ uniquely identifies matched segments, while $l < 0.5$ enables the incorporation of partial overlaps due to segmentation errors. When both thresholds are fixed at 0.5, \softpq\ reduces to the original PQ metric. It is the interval between these two thresholds that constitute the "soft" aspect of \softpq\ and this softness is adjustable. This makes \softpq\ a convenient generalization as it ensures backward compatibility while offering additional flexibility for more nuanced evaluations.

\begin{table}[ht]
\centering
\begin{tabular}{p{0.95\linewidth}}
\hline 
\textit{\textbf{Algorithm: Modified Panoptic Quality}} (SoftPQ) \\ \hline
\textbf{Input:} \\
\(P\): Set of non-overlapping predicted segments, $p$ \\
\(G\): Set of ground truth segments, $g$ \\
\(l\): Lower IoU threshold \\
\(h\): Upper IoU threshold \\
\(\nsoft_g\): Number of soft-match segments for a given ground-truth\\ 
\(\nmatch_g\): Number of true positives \\ \hline 
\textbf{Output:} \\
SoftPQ: Modified Panoptic Quality metric \\ \hline
\textbf{Steps (for evaluating an image or image set):} \\
1. Initialize: \\
\hspace*{1em} \(\IoUh \gets 0\) \\
\hspace*{1em} \(\IoUl \gets 0\) \\

2. For each predicted segment \(p \in P\): \\
\hspace*{1em} For each ground truth segment \(g \in G\): \\
\hspace*{2em} Compute \(\mathrm{IoU}_{p, g}\) \\
\hspace*{2em} If \(\mathrm{IoU}_{p, g} \geq h\): \\
\hspace*{3em} Mark \(p\) and \(g\) as match \\
\hspace*{3em} \(\IoUh \gets \IoUh + \mathrm{IoU}_{p, g}\) \\
\hspace*{2em} Else if \(l < \mathrm{IoU}_{p, g} < h\): \\
\hspace*{3em} Mark \(p\) as soft-match \\
\hspace*{3em} \(\IoUl \gets \IoUl + \mathrm{IoU}_{p, g}\) \\

3. Compute Modified IoU: \\
\hspace*{1em} \(\IoUmod \gets \IoUh + \frac{\IoUl}{\sqrt{\nsoft + 1}}\) \\

4. Compute Modified $\mathrm{SoftPQ}$: \\
\hspace*{1em} If \(\nmatch = 0\): \\
\hspace*{2em} \(\mathrm{SoftPQ} \gets \frac{\IoUmod}{\GTcount}\) \\
\hspace*{1em} Else: \\
\hspace*{2em} \(\mathrm{SoftPQ} \gets \frac{\IoUmod \times F_1}{\nmatch}\) \\

5. Return $\mathrm{SoftPQ}$ \\ \hline
\end{tabular}
\label{tab:modified_pq_pseudocode}
\end{table}

\subsection{Modified IoU Calculation}

To provide a more versatile assessment of segmentation quality, especially in cases involving partial or near-miss predictions, we introduce a modified IoU formulation that incorporates both strong and weak matches. Specifically, we separate predicted segments into two categories based on their IoU with ground truth segments: those that exceed a high-confidence threshold $h$ (i.e., $\mathrm{IoU} \geq h$) and those that fall between a lower bound $l$ and the upper threshold (i.e., $l < \mathrm{IoU} < h$). The former contribute to a cumulative sum denoted as $\IoUh$, while the latter are aggregated into a partial score $\IoUl$. Let $\nsoft_g$ be the number of soft-match predictions $p \in P$ in this partial range for a given ground-truth segment $g \in G$.
The modified IoU is then defined as:

\begin{align}
    \IoUmod_g &\equiv \IoUh_g + \frac{1}{\sqrt{\nsoft_g + 1}}\, \IoUl_g    \label{eq:iou_modified}  \\
   & = \sum_{p\in P} \left[ \mathds{1}_{\mathrm{IoU}_{p,g} \geq h}  +  \frac{ \mathds{1}_{l < \mathrm{IoU}_{p,g} < h}}{\sqrt{\nsoft_g+1}}  \right]   \mathrm{IoU}_{p,g} 
\end{align}

where
\begin{equation}
    \nsoft_g = \sum_{p \in P} \mathds{1}_{l < \mathrm{IoU}_{p,g} < h}
\end{equation}
is the number of softmatch segments for a given $g$, and $\mathds{1}_{X}$ is the indicator function, zero when its argument $X$ is false and unity when true.

This is recognized as a weighted sum of the matched and softmatched predictions. The $\sqrt{\nsoft_g + 1}$ denominator ensures that as the number of partial (soft) matches increases, their influence on the final score grows more slowly, preventing large collections of weak overlaps from disproportionately inflating the metric. At the same time, it preserves sensitivity to incremental improvements by allowing near-miss predictions to contribute in a scaled manner. When $\nsoft_g = 0$, the expression reduces to a sum over confidently matched segments, maintaining consistency with the standard PQ formulation.

More broadly, this approach can be interpreted as a specific instance of a general aggregation framework:
\begin{equation}
   \IoUmod = \IoUh + f(\nsoft)\, \IoUl \label{eq:iou_general}
\end{equation}
where $f(\nsoft)$ is a monotonic weighting function that moderates the contribution of uncertain matches. In this work, we select $f(\nsoft) = 1/\sqrt{\nsoft + 1}$ based on empirical performance and its more gradual effect, though other choices are possible. Further discussion regarding this can be found in the Appendix~\ref{sec:ablation_penalty}.

\subsection{Modified PQ Metric}
The final modified Panoptic Quality metric (\( \SoftPQlh \)) incorporates the precision-recall trade-off using the \( F_1 \)-score and is calculated differently depending on the presence of true positives (\( \nmatch \)):

\begin{itemize}
    \item \textbf{No True Positives (\( \nmatch = 0 \))}: If no predictions are matched, the metric is normalized by the total number of ground truth segments (\( \GTcount \)):
    \begin{equation}
        \SoftPQlh = \frac{\IoUmod}{\GTcount}
    \end{equation}
    \item \textbf{With True Positives (\( \nmatch > 0 \))}: If matches exist, the Modified IoU is scaled by the \( F_1 \)-score and normalized by the number of true positives (\( \nmatch \)):
    \begin{equation}
        \SoftPQlh = \frac{\IoUmod \, \, F_1}{\nmatch}
    \end{equation}
\end{itemize}

By recalibrating the traditional Panoptic Quality metric with adjustable thresholds and a nuanced IoU computation, the \softpq\ provides a comprehensive and flexible framework for segmentation evaluation.

\subsection{Controlled Sensitivity for under- and over-segmentation}
As the lower IoU threshold is used to measure soft match between the true and predicted labels, this can be used to either detect under-segmentation or over-segmentation. Neither is ideal, but both are common in practice. Depending on the experiment and application sensitivity, one can be prioritized over other. Proposed method, allows the user to prioritize depending on the task at hand. For instance, in applications where under-segmentation (merging distinct objects) is more detrimental than over-segmentation (splitting objects), the thresholds can be tuned to penalize such errors more strongly. This enables task-specific metric modification, making \softpq\ adaptable to a wide range of domains. The equations 1-3 and pseudo code given above prioritizes over-segmentation over under-segmentation. For under-segmentation, the equations can be modified as below:

\begin{align}
    \IoUmod_p &\equiv \IoUh_p + \frac{1}{\sqrt{\nsoft_p + 1}}\, \IoUl_p    \label{eq:iou_modified_2}  \\
   & = \sum_{g\in G} \left[ \mathds{1}_{\mathrm{IoU}_{p,g} \geq h}  +  \frac{ \mathds{1}_{l < \mathrm{IoU}_{p,g} < h}}{\sqrt{\nsoft_p+1}}  \right]   \mathrm{IoU}_{p,g} 
\end{align}

where
\begin{equation}
    \nsoft_p = \sum_{g \in G} \mathds{1}_{l < \mathrm{IoU}_{p,g} < h}
\end{equation}
is now the number of softmatch segments for a given $p$, and $\mathds{1}_{X}$ is still the indicator function.

Example images and code will be made publicly available on our GitHub\footnote{https://github.com/rkarmaka/SoftPQ}.

\section{Experiments}

We evaluate the proposed metric, \softpq\, through a series of controlled experiments designed to assess its sensitivity, robustness, and interpretability in comparison to traditional segmentation metrics such as F1, mAP, IoU, and PQ.

\subsection{Sensitivity to Progressive Erosion}

\begin{figure}[h!]
    \centering
    \includegraphics[width=\linewidth]{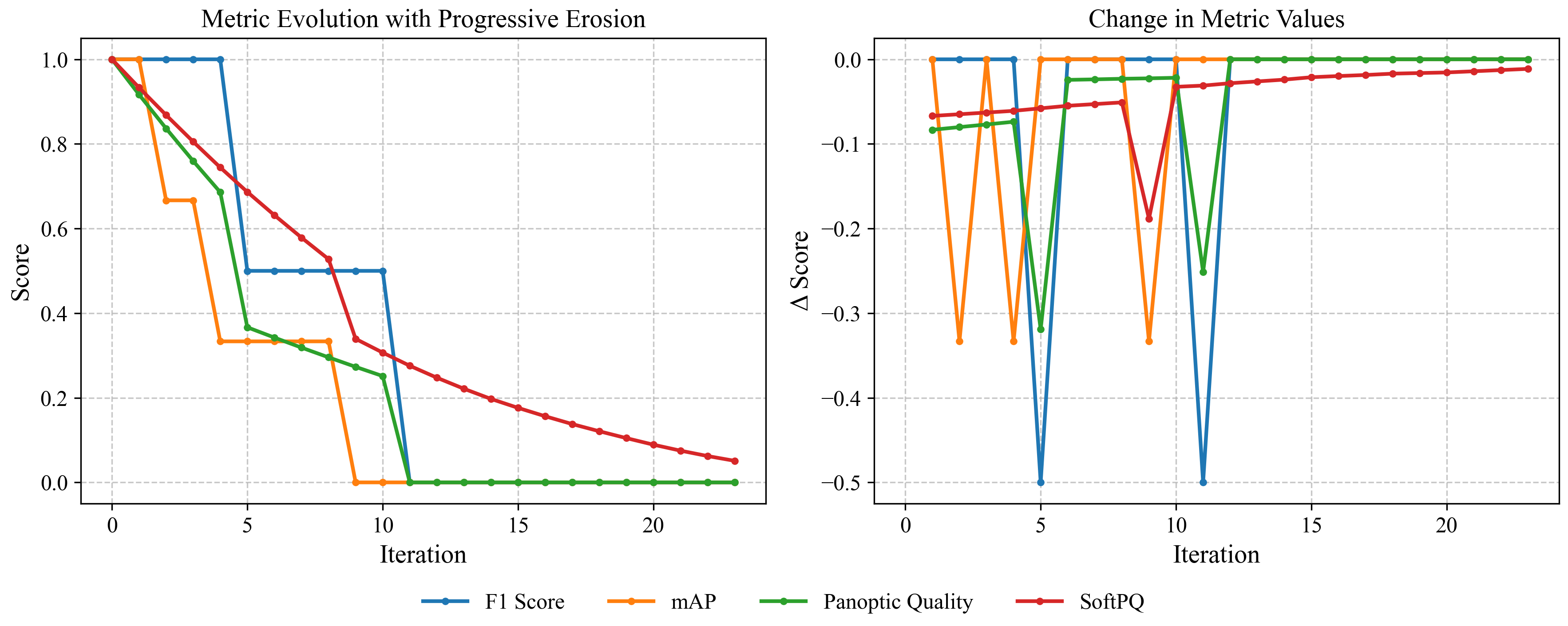}
    \caption{\textbf{Metric Sensitivity to Progressive Erosion:} (Left) Metric scores as predicted masks are eroded over $25$ iterations, simulating increasing under-segmentation. (Right) Per-step score change ($\Delta$Score) highlights the stability and sensitivity of each metric. \softpq\ (red) shows a smooth, consistent decline, while F1, mAP, and PQ respond abruptly or inconsistently to minor perturbations. This illustrates \softpq's superior resolution in capturing gradual degradation.}
    \label{fig:delta_erosion_plot}
\end{figure}

To evaluate the sensitivity of different metrics to gradual under-segmentation, we iteratively erode the predicted masks and track how the scores evolve across 20 steps. As shown in Figure~\ref{fig:delta_erosion_plot}, the left panel plots the raw score trajectories, while the right panel visualizes the change in score between successive iterations.

\softpq\ exhibits the smoothest and most consistent decline, enabled by its use of a low IoU threshold and sublinear penalty function. In contrast, traditional metrics such as F1 and mAP display sharp, discrete drops or periods of insensitivity, failing to reflect fine-grained changes in mask quality. PQ also suffers from step-like behavior due to its hard matching criteria. These results highlight \softpq's value in iterative optimization and model debugging workflows, where developers require steady and interpretable feedback on small segmentation refinements.

\subsection{Sensitivity to IoU Threshold Sweeping}

We examine how \softpq\ behaves under varying degrees of match leniency by sweeping the lower IoU threshold $l$ from 0.05 to 0.5, while fixing the upper threshold $h=0.5$. This interpolation controls the softness of the matching criteria, allowing \softpq\ to assign partial credit for predictions that fall short of strict overlap.

\begin{figure}[h!]
    \centering
    \includegraphics[width=\linewidth]{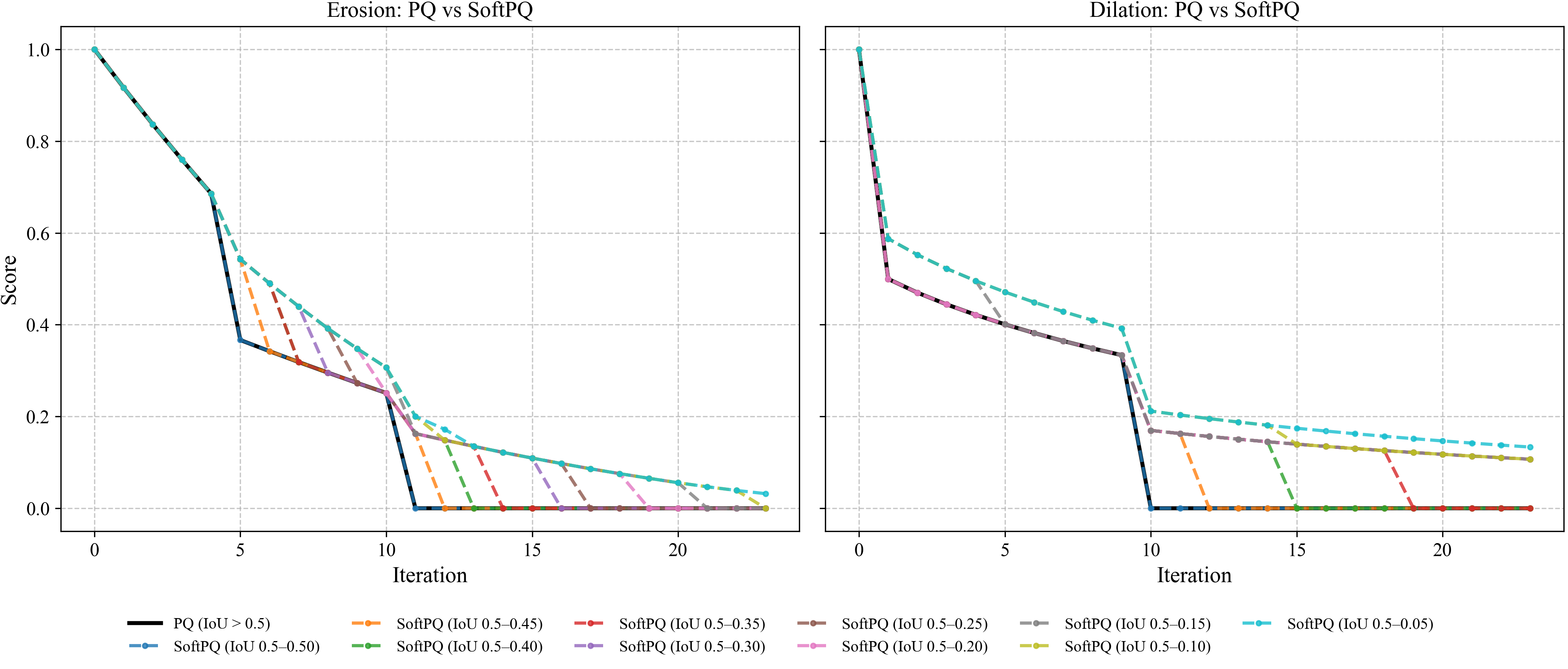}
    \caption{\textbf{Sensitivity of \softpq\ to Lower IoU Thresholds:} Left: Progressive dilation; Right: Progressive erosion of predicted masks. Each curve represents \softpq\ computed with a different lower IoU threshold $l$, while fixing the upper threshold $h=0.5$. As thresholds become more lenient, \softpq\ scores degrade more smoothly and maintain interpretability under noisy or degraded predictions. In contrast, PQ (black) drops sharply when predictions fall below its strict matching condition, failing to reflect gradual changes in quality.}
    \label{fig:iou_sweep}
\end{figure}

Figure~\ref{fig:iou_sweep} shows two complementary cases: progressive dilation (left) and erosion (right) of predicted masks. In both scenarios, \softpq\ reveals a smooth gradation of score degradation as segmentation quality decreases. Lower thresholds (e.g., $l=0.5$) provide a more forgiving evaluation, maintaining non-zero scores even under significant mask perturbation. In contrast, the traditional PQ metric (black line) collapses abruptly once predictions fall below the rigid $\mathrm{IoU} > 0.5$ threshold.

This controlled sweeping demonstrates that \softpq\ enables flexible, application-aware evaluation, revealing partial improvements and offering more interpretable feedback during model development.

\subsection{Robustness to over-segmentation}

\begin{figure}[ht!]
    \centering
    \includegraphics[width=\linewidth]{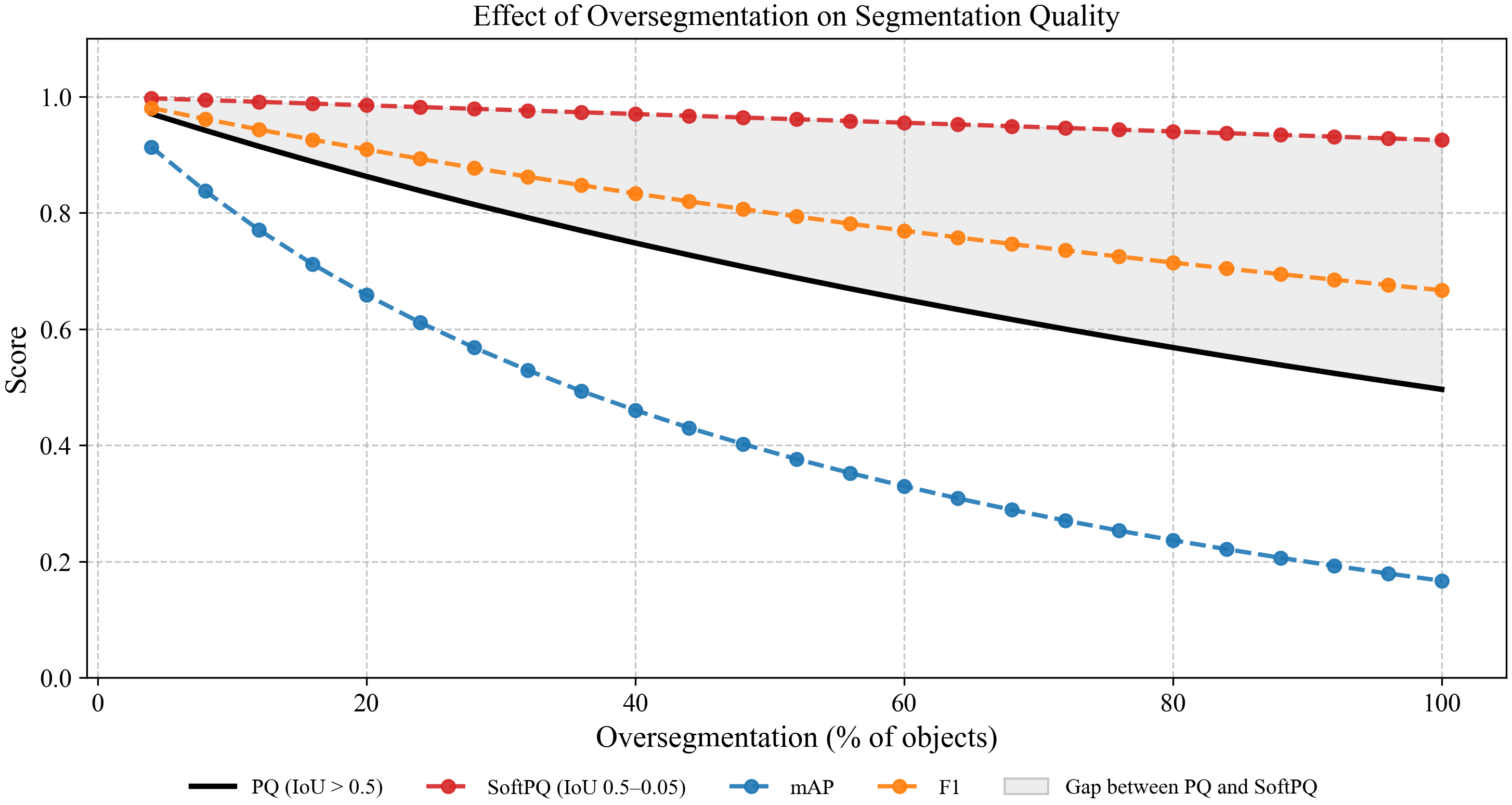}
    \caption{\textbf{Robustness to over-segmentation:} Impact of over-segmentation on various segmentation metrics. Each ground truth object is progressively split into 1–5 predicted segments, while maintaining a perfect 1:1 object-level match. \softpq\ (red) shows the most robust behavior, degrading gradually due to its tolerance for partial matches at low IoU thresholds ($l=0.05$). PQ (black) exhibits a more rigid drop, and mAP (blue) declines sharply due to penalizing redundant detections. The shaded region represents the envelope of \softpq\ scores across different $l$ values, indicating the flexibility of the metric under controlled segmentation fragmentation.}
    \label{fig:overseg_plot}
\end{figure}

We simulate over-segmentation by progressively splitting each of 25 ground truth objects into multiple predicted segments, ranging from no splits to five fragments per object. Importantly, despite the fragmentation, each object is still uniquely detected, maintaining a 1:1 correspondence between ground truth and predicted instances. Figure~\ref{fig:overseg_plot} illustrates how different metrics respond to this controlled perturbation.

\softpq\ demonstrates the highest robustness to over-segmentation, with only a modest decline in score even at 100\% over-segmentation. This contrasts sharply with mAP, which drops significantly due to its penalization of redundant detections. PQ, while more stable than mAP, still fails to account for the nuanced partial correctness of fragmented predictions. The gray band in Figure~\ref{fig:overseg_plot} visualizes the region spanned by \softpq\ at varying lower IoU thresholds, with $h=0.5$ fixed. As the lower bound $l$ approaches zero, \softpq\ captures the latent correctness that PQ misses, highlighting its capacity to gracefully handle structured segmentation errors.

\section{Results and Discussion}
\label{sec:results}
We report findings from a series of controlled segmentation experiments designed to assess the behavior of \softpq\ relative to standard metrics such as F1, mAP, IoU, and PQ. The results demonstrate that \softpq\ offers smoother, more interpretable feedback across a range of conditions, providing both robustness to structural noise and flexibility for model developers.

\subsection{Differentiating Error Types with Gradual Penalties}

Conventional metrics often fail to distinguish between qualitatively different segmentation errors. For example, F1 and PQ may assign similar scores to both mildly and severely oversegmented predictions. In contrast, \softpq\ incorporates a soft matching regime and a configurable penalty function that enables more expressive scoring. It assigns higher scores to predictions that exhibit near-misses—such as slightly shifted masks or minor over-segmentations—while still penalizing severe errors like ghost detections or incorrect merges. This nuanced behavior helps developers identify whether errors are catastrophic or recoverable.

\subsection{Smooth and Interpretable Response to IoU Threshold Variation}

The flexibility of \softpq\ is most clearly demonstrated through IoU threshold sweeping. By varying the lower threshold $l$ from 0.01 to 0.5 while fixing the upper threshold $h = 0.5$, \softpq\ reveals a smooth, monotonic score curve that interpolates between highly lenient and strict matching conditions. This property, illustrated in Figure~\ref{fig:iou_sweep}, makes \softpq\ a powerful tool for visualizing partial improvements that standard PQ fails to capture. This granularity is useful during early training or with noisy ground truth.

\subsection{Robustness to over-segmentation}

To test resilience under structured over-segmentation, we progressively split each of 25 ground truth objects into multiple predicted segments—up to five per object—while maintaining a perfect 1:1 object-level match. As shown in Figure~\ref{fig:overseg_plot}, mAP degrades rapidly due to redundant detections, and PQ also declines steadily. In contrast, \softpq\ remains significantly more stable, thanks to its tolerance for low-IoU partial matches. The shaded region in the plot illustrates the envelope of \softpq\ values under varying lower thresholds, emphasizing its flexibility in reflecting consistent partial correctness even when predictions are fragmented.

\subsection{Real-World Comparison: PQ vs. SoftPQ on Cellpose Dataset}
\begin{figure}[ht!]
\centering
\includegraphics[width=\linewidth]{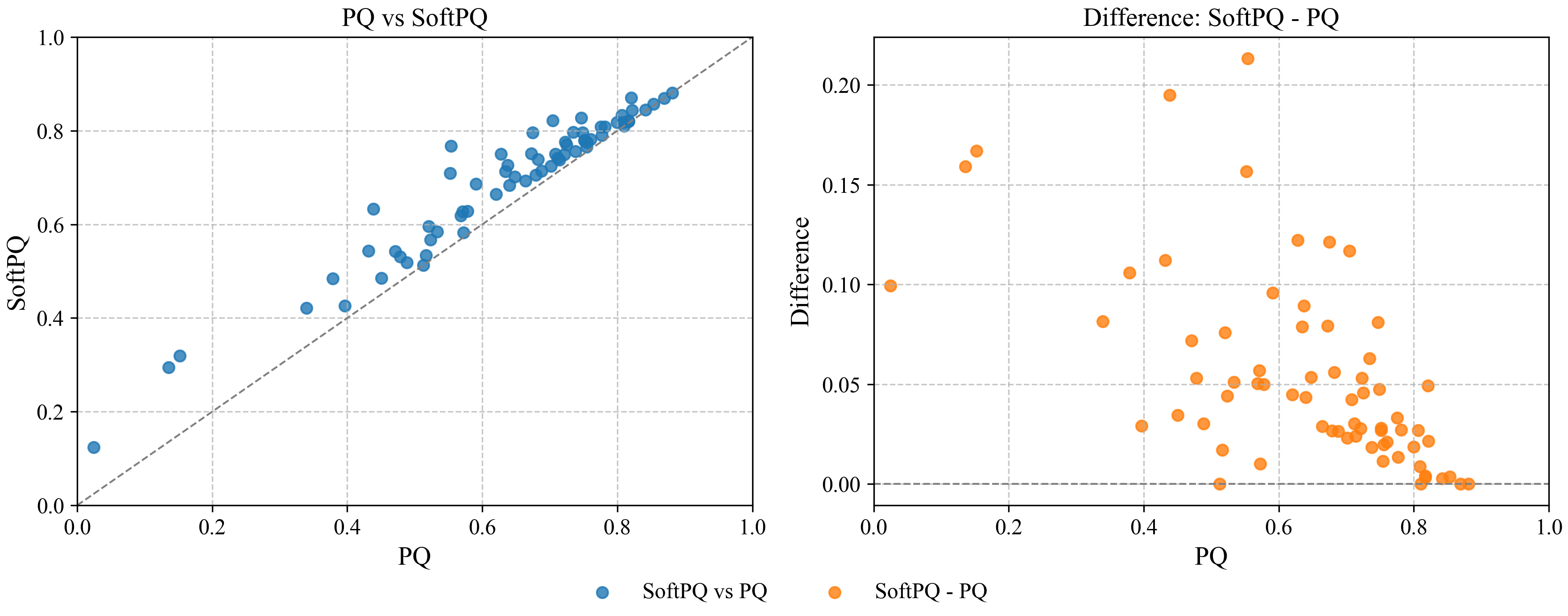}
\caption{\textbf{Comparison between PQ and \softpq\ on the Cellpose dataset:} (Left) \softpq\ scores plotted against PQ for each sample. \softpq\ is equal to or higher than PQ in all cases. (Right) The difference \softpq\ -- PQ plotted against PQ. Larger improvements from \softpq\ are observed in lower-PQ cases, often associated with over-segmentation or under-segmentation errors.}
\label{fig:pq_vs_softpq}
\end{figure}

To evaluate how \softpq\ performs in practical settings, we applied both the standard PQ and the proposed \softpq\ ($l = 0.05$, $h = 0.5$) to real-world cell segmentation outputs from the Cellpose dataset~\cite{stringer2021cellpose}. In the left panel of Figure~\ref{fig:pq_vs_softpq}, we compare PQ and \softpq\ scores directly for each image. Nearly all data points lie on or above the identity line, indicating that \softpq\ returns scores equal to or higher than PQ. This confirms that when predictions are neither oversegmented nor undersegmented, both metrics agree. However, in cases involving segmentation errors, particularly splits or merges, \softpq\ captures additional soft matches that PQ overlooks due to its stricter thresholding.

The right panel highlights the difference (\softpq\ -- PQ) as a function of PQ. Larger differences occur primarily when the PQ score is lower, which corresponds to images with more complex segmentation errors. This shows the robustness of \softpq\ where PQ may fail to capture useful information.

\subsection{Guidelines for Choosing Thresholds in Real-World Settings}

In practice, the choice of the thresholds should be guided by the nature of the application and the types of errors that are most critical to penalize. For general-purpose benchmarking, we find that setting $h = 0.5$ and $l = 0.25$ strikes a useful balance, offering meaningful differentiation without overreacting to small perturbations.

In high-precision domains such as medical imaging or safety-critical manufacturing, stricter thresholds (e.g., $h = 0.6$–$0.75$, $l = 0.4$–$0.5$) are advisable, as they ensure that only highly accurate predictions contribute significantly to the score. Conversely, in noisy environments or during early model development, where segmentation outputs may be imperfect, more relaxed thresholds (e.g., $l = 0.05$–$0.15$) allow \softpq\ to reflect gradual improvements and maintain useful feedback.

Moreover, the relative impact of over- and under-segmentation can be adjusted via the choice of $l$. Increasing $l$ penalizes under-segmentation more heavily by making it harder for merged segments to earn partial credit. We recommend that practitioners explore small sweeps of these parameters on validation sets to identify a configuration that aligns with task priorities and downstream use cases.

\section{Limitations}
Similar to PQ, a key limitation of this work is the assumption that segmentation outputs do not contain overlapping masks. In practical applications, particularly in complex scenes with occluded, overlapping segments can naturally arise. Some segmentation methods may produce masks that partially overlap for valid reasons, and \softpq\ in its current form does not explicitly account for this scenario.

\section{Conclusion}

We introduced \softpq\, a generalized extension of the PQ metric that incorporates soft matching to provide a more flexible and interpretable framework for segmentation evaluation. Unlike traditional metrics that rely on fixed IoU thresholds and binary decisions, 	\softpq\ defines correctness as a continuum, using tunable upper and lower thresholds to capture partial overlaps and ambiguous predictions.

This soft-match formulation is the core innovation of \softpq. By relaxing the rigid matching condition and introducing a sublinear penalty function, \softpq\ is able to differentiate between severe and minor segmentation errors. It rewards gradual improvements and provides smoother feedback during model development, which is often obscured by conventional metrics such as F1, IoU, and PQ. Critically, \softpq\ reduces to the original PQ when both thresholds are set to 0.5, ensuring backward compatibility. This makes \softpq\ a strict generalization rather than a replacement. It allows users to calibrate the metric based on task-specific sensitivity to errors like over- and under-segmentation, adapting the evaluation process to better match application needs.

Through a set of controlled experiments, we demonstrated that \softpq\ offers more stable behavior under progressive degradation, better captures qualitative differences in predictions, and enables finer control over penalty behavior. These properties make \softpq\ a practical and principled alternative for both benchmarking and iterative model refinement. We hope \softpq\ provides a foundation for more adaptable and meaningful evaluation practices in instance segmentation, and encourages further exploration into soft-matching strategies for other structured prediction tasks.

\bibliographystyle{plain}


\newpage
\appendix

\section{Technical Appendices and Supplementary Material}

\subsection{Ablation of Penalty Function $f(\nsoft)$} \label{sec:ablation_penalty}

To analyze the effect of different penalty functions in \softpq\, we evaluate three variants: linear ($f^{-1}(\nsoft) = n+1$), mildly sublinear ($f^{-1}(\nsoft) = \sqrt{n+1}$), and strongly sublinear ($f^{-1}(\nsoft) = \log(\nsoft+1)$) scaling. These functions modulate the penalty applied when multiple overlapping segments match a single ground truth object, effectively controlling how fragmentation impacts the final score.

\begin{figure}[ht!]
\centering
\includegraphics[width=\linewidth]{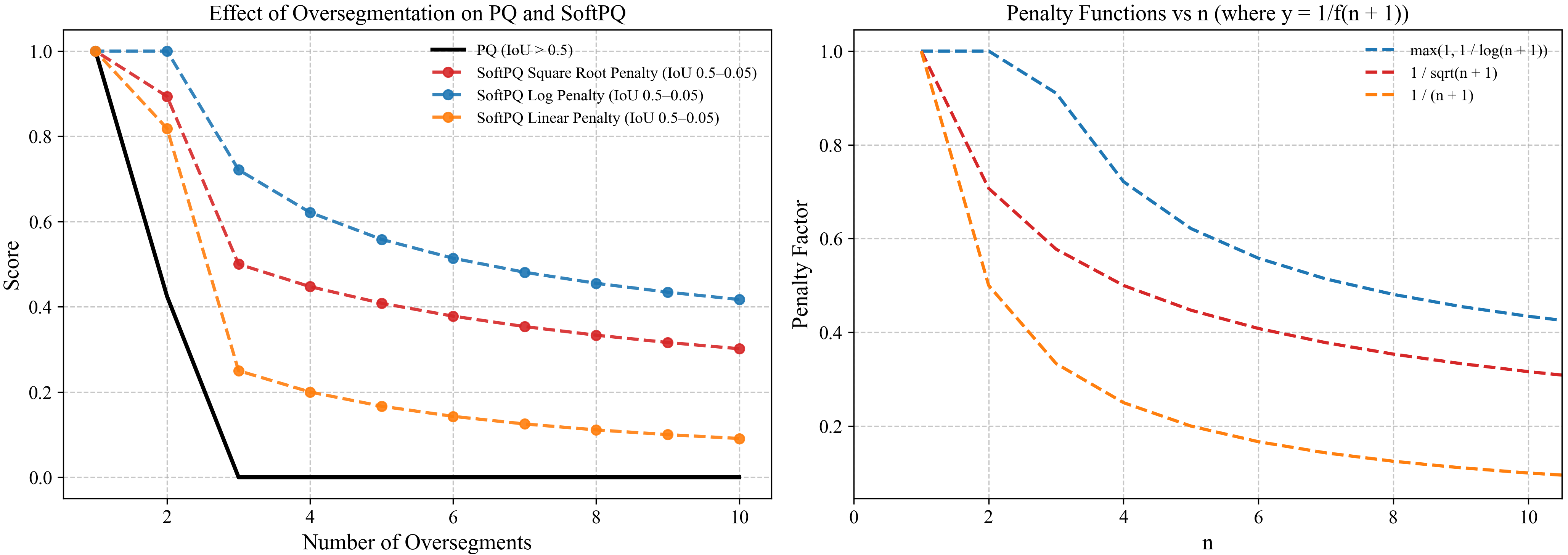}
\caption{\textbf{Effect of Penalty Function $f(\nsoft)$ on 	\softpq\ Behavior:} (Left) Comparison of \softpq\ performance under three penalty functions—linear ($f^{-1}(\nsoft) = n+1$), sublinear ($f^{-1}(\nsoft) = \sqrt{n+1}$), and logarithmic ($f^{-1}(\nsoft) = \log(\nsoft+1)$)—as the number of predicted fragments per object increases. Sublinear scaling (red) provides the most stable and gradual decline in score, while linear and logarithmic functions produce steeper penalties for over-segmentation. (Right) Visualization of the inverse scaling values ($f^{-1}(\nsoft)$) used in \softpq\ weighting, illustrating the analytical behavior of each function.}
\label{fig:penalty_plot}
\end{figure}

As illustrated in Figure~\ref{fig:penalty_plot}, the right panel shows the mathematical behavior of these functions, while the left panel demonstrates their practical impact in a controlled over-segmentation scenario. Both linear and logarithmic penalties react sharply to fragmentation, resulting in steep drops in \softpq\ scores. In contrast, the sublinear function yields a more gradual decline, offering greater stability and robustness. This behavior better preserves distinctions between minor and major segmentation errors, making sublinear penalties preferable in settings that demand consistent tolerance to noise.

The design of $f(\nsoft)$ in \softpq\ thus enables tunable behavior: linear and logarithmic penalties suit high-precision applications requiring strict error handling, while sublinear penalties are well-suited for exploratory or clinical domains where interpretability and robustness are prioritized.

\end{document}